\documentclass[10pt,journal,compsoc]{IEEEtran}
\usepackage{amsmath}
\usepackage{amssymb}
\usepackage{algorithm}
\usepackage{graphics}
\usepackage{graphicx}
\usepackage{tabularx}
\usepackage[noend]{algpseudocode}
\usepackage{multirow}
\usepackage{arydshln}
\usepackage{booktabs}  
\ifCLASSOPTIONcompsoc
  \usepackage[nocompress]{cite}
\else
  \usepackage{cite}
\fi
\usepackage{mathtools}
\DeclarePairedDelimiterX{\norm}[1]{\lVert}{\rVert}{#1}
\ifCLASSINFOpdf
\else
\fi

\usepackage{xcolor}

\newcommand{\shadow}[1]{}
\newcommand{\blue}[1]{\textcolor{black}{#1}}

\def\b{\blue}
\def\s{\shadow}

\begin{document}
%
\title{DASVDD: Deep Autoencoding Support Vector Data Descriptor for Anomaly Detection}
%
%
%
%

\author{Hadi~Hojjati,
        and~Narges~Armanfard
\IEEEcompsocitemizethanks{\IEEEcompsocthanksitem Both authors are with the Department of Electrical and Computer Engineering, McGill University, and Mila - Quebec AI Institute, Montreal, QC, Canada.\protect\\
E-mail: hadi.hojjati@mcgill.ca\\narges.armanfard@mcgill.ca
}
}

%
%

\markboth{IEEE Transaction on Knowledge and Data Engineering}%
{Hojjati and Armanfard: Deep Autoencoding Support Vector Data Descriptor for Anomaly Detection}
%



\IEEEtitleabstractindextext{%
\begin{abstract}
One-class anomaly detection aims to detect anomalies from normal samples using a model trained on normal data. With recent advancements in deep learning, researchers have designed efficient one-class anomaly detection methods. Existing works commonly use neural networks to map the data into a more informative representation and then apply an anomaly detection algorithm. In this paper, we propose a method, DASVDD, that jointly learns the parameters of an autoencoder while minimizing the volume of an enclosing hypersphere on its latent representation. We propose a novel anomaly score that combines the autoencoder's reconstruction error and the distance from the center of the enclosing hypersphere in the latent representation. Minimizing this anomaly score aids us in learning the underlying distribution of the normal class during training. Including the reconstruction error in the anomaly score ensures that DASVDD does not suffer from the hypersphere collapse issue since the DASVDD model does not converge to the trivial solution of mapping all inputs to a constant point in the latent representation. Experimental evaluations on several benchmark datasets show that the proposed method outperforms the commonly used state-of-the-art anomaly detection algorithms while maintaining robust performance across different anomaly classes.
\end{abstract}

}

\maketitle

\IEEEdisplaynontitleabstractindextext

%
\IEEEpeerreviewmaketitle

\IEEEraisesectionheading{\section{Introduction}\label{intro}}

%
%
%
%
\IEEEPARstart{A}{nomaly} detection (AD) is the task of identifying samples of a dataset that deviate from the "normal" pattern \cite{10.1145/1541880.1541882} (the term "normal" is unrelated to the Gaussian distribution here and elsewhere in the paper, unless otherwise specified). Anomaly detection has been an active field of research in recent years due to its application in a wide variety of domains, including fraud detection \cite{creditsurvey}, medical care \cite{Gugulothu2018SparseNN}, time-series anomaly detection \cite{Buda2018DeepADAG}, video surveillance \cite{kiran2018overview}, machine vision applications \cite{golan2018deep}, and industrial monitoring \cite{doi:10.1177/1475921717737051, hojjati}.
Typically, in most AD problems, we are given samples from the normal class\b{, and the goal is to train a model that describes the pattern of the given data and thus identify the instances that deviate from the normal pattern as anomalies \cite{pang}}. This procedure is also known as one-class classification.
A long line of literature, such as one-class support vector machines (OCSVMs) \cite{ocsvm}, kernel density estimation (KDE) \cite{kde}, and more recently, Isolation Forests (IFs) \cite{IF}, has addressed \b{this} problem with classical (i.e., non-deep learning) machine learning methods.

\begin{figure*}[t]
\label{fig: overview}
  \centering
  \includegraphics[width=1\linewidth]{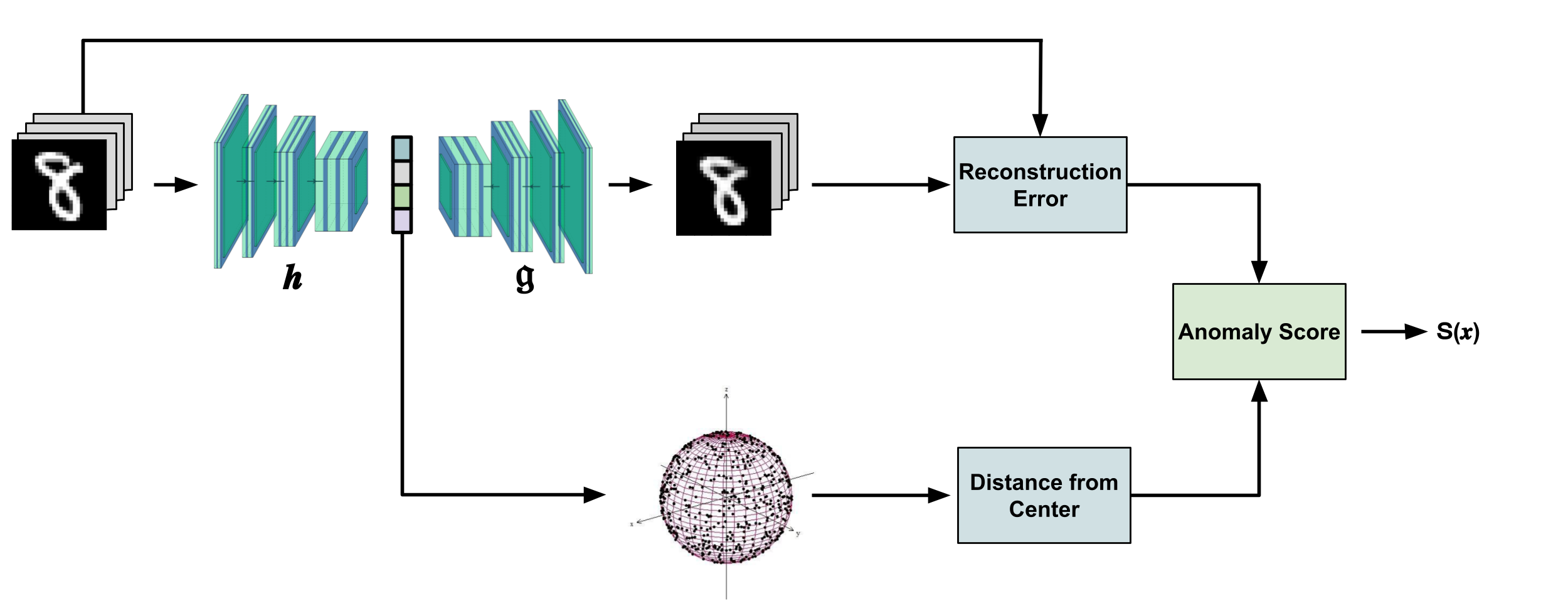}
  \caption{Overview of the proposed method DASVDD.}
  \label{fig:overview}
\end{figure*}

In the past decade, deep anomaly detection algorithms have become popular thanks to the development of Autoencoders (AEs) \cite{ae,ae2} and Generative Adversarial Networks (GANs) \cite{gansurvey}. Deep AD methods can be divided into two categories: the first group of algorithms train an AE or GAN on normal data and uses the reconstruction error of new samples for identifying anomalies \cite{sakurada}. The second group uses neural networks to extract a lower-dimensional representation of input data and feed it to a classical AD algorithm such as OCSVM \cite{sabokrou2017deepanomaly}.

\b{Another line of research in deep anomaly detection focuses on developing domain-specific AD algorithms. This research area has attracted a lot of attention from the community, especially from computer vision researchers, and has significantly outperformed the results of other methods \cite{golan2018deep,tack2020csi}. While they have achieved unprecedented success in image anomaly detection, these models commonly use a domain-specific transformation in their pipeline, which makes them unsuitable for other input types \cite{hojjati2022selfsupervised}.}

\b{Recently, Ruff et al. \cite{ruff2018deep} proposed a deep anomaly detection method called Deep Support Vector Data Descriptor (DSVDD)}. In contrast to other state-of-the-art domain-specific models, DSVDD is suitable for general AD and does not depend on the input type. \b{It combines two widely-used anomaly detection approaches: training a single neural network for simultaneously extracting a lower-dimensional representation of data and a support vector data descriptor (SVDD) that minimizes the volume of the enclosing hypersphere on this lower-dimensional representation.} \blue{DSVDD has emerged as an efficient anomaly detection algorithm since it is an AD algorithm that, unlike self-supervised methods, does rely on input-specific transformations and is applicable to a wide range of input types. Among the non-self-supervised models, DSVDD has been shown to outperform most baselines \cite{ruff2018deep}. As a result, many anomaly detection researchers have developed methods inspired by DSVDD.} Although DSVDD is a promising algorithm and produces encouraging outcomes, it suffers from a vital problem during training which is called ``hypersphere collapse'' \cite{ruff2018deep, maziarka2020flowbased}. Hypersphere collapse occurs when the network converges to the trivial solution of all-zero weights (this phenomenon will be discussed in more detail in the next section). To prevent this problem, authors of DSVDD \cite{ruff2018deep} imposed several constraints on the architecture of their deep network, such as fixing the hypersphere center and setting network biases to zero, which limits the performance and effectiveness of the algorithm \cite{maziarka2020flowbased}.

In this paper, we propose a novel anomaly detection algorithm inspired by DSVDD and Autoencoders. Our method, deep autoencoding support vector data descriptor (DASVDD), trains an autoencoder instead of a vanilla neural network by simultaneously minimizing the volume of the enclosing hypersphere in the learned latent representation of the encoder and the reconstruction error of the decoder's output. Therefore, the autoencoder learns to map the normal data into a hypersphere with a minimum volume while can still reconstruct the original input. We propose a customized anomaly score which is a combination of the AE's reconstruction error and the distance of the sample from the center of the hypersphere. The block diagram of the proposed DASVDD is shown in Figure \ref{fig:overview}. We show that our model does not encounter the hypersphere collapse problem even though it considers the hypersphere center and network biases as parameters that are trained during the training phase. We present extensive experiments of DASVDD on the benchmark datasets and compare its performance against state-of-the-art algorithms. The benchmark datasets include \b{three computer vision datasets and \blue{four} datasets from other domains, including speech, industrial sounds, intrusion detection, and medical data}. The effective performance of DASVDD on these datasets shows the robustness of the proposed method across a wide variety of anomaly detection applications.

In summary, the main contributions of our paper are as follows:
\begin{enumerate}
    \item \b{We propose an SVDD-based anomaly detection approach which prevents the hypersphere collapse problem that other deep SVDD algorithms face.} We propose to use the autoencoder reconstruction error as a term in the loss function. This prevents the trivial all-zero solution for the parameters.
    
    \item \b{In our model, unlike other SVDD-based deep models, the hypersphere center is a free optimization parameter and can be optimized during training. These methods, such as DSVDD \cite{ruff2018deep}}, commonly fix the hypersphere center prior to the training to prevent the network from converging to the trivial solution. This may lead to a suboptimal solution because of forcing the hypersphere center to be equal to a predefined value.
    
    \item We propose an iterative training strategy for jointly minimizing the network's reconstruction loss and the DSVDD loss. Our strategy prevents the total loss function from collapsing to either of the mentioned losses and provides a stable training procedure.
    
    \item Our approach requires a hyperparameter which balances the contribution of the two loss functions. Finding an appropriate value for this parameter is challenging because it might lead to the model's collapse to either of the two loss terms. In this paper, we also propose an effective approach for tuning this hyperparameter prior to the training and experimentally show that the resulting value leads to a stable training procedure.
    
    \item Unlike domain-specific models, our proposed approach does not use any \b{input-related} transformations and is thus suitable for general anomaly detection. \b{We test our model on different types of datasets, including images, speech, biomedical data, network data, and industrial acoustic recordings to show its efficiency across various applications.}
\end{enumerate}

The remainder of this paper is organized as follows: In section 2, we review the state-of-art algorithms in anomaly detection, with a focus on Deep SVDD-based methods. In the third section, we describe our proposed method, DASVDD, and its properties. Section 4 presents an extensive evaluation of DASVDD on several benchmark datasets. We also discuss some of the properties and possible expansions of our model. Finally, section 5 offers some concluding remarks.

\section{Related Works}
In this section, we briefly review the existing methods in deep anomaly detection. One can refer to \cite{10.1145/1541880.1541882,chalapathy2019deep} for a recent and detailed review of past literature in the AD field.

Deep autoencoders \cite{Hinton504} are some of the most commonly-used deep models for anomaly detection. AEs can be used for extracting a lower-dimensional representation of the data while retaining their common factors of variation. AE-based AD algorithms either feed this latent representation to a conventional anomaly detection algorithm, such as Gaussian Mixture Model (GMM) or one-class SVM (OCSVM) \cite{erfani} or directly use the reconstruction error as their anomaly measure \cite{vae}\cite{cae}.
Several variants of autoencoders, including variational AEs, \cite{vae}, and convolutional AEs \cite{cae}, have been used for anomaly detection in different applications \cite{vaeapplication}. These methods assume that the autoencoder can extract the common factors of variation from normal data and thus can reconstruct them with fair accuracy, but \b{they fail to do so with} anomalies because they do not possess the  common factors associated with the normal data.
This assumption, however, does not always hold. In some cases, an autoencoder which is trained on the normal data can also reconstruct anomalies accurately \cite{gong2019memorizing}. To prevent this problem, it is important to adopt an appropriate network architecture and latent representation size. Choosing the right size for the latent space often depends on the data and the task that we would like to carry out. Since we usually do not have access to all anomaly classes, it is tough to find an optimal size for the bottleneck of the AE network.

Recently, several studies have developed AD algorithms based on GANs \cite{dimattia2019survey}. In most GAN-based approaches, the goal is to train the latent representation of the generator network so that it captures the underlying pattern of the normal data. These methods assume that the generative network can generate normal instances from the latent space better than anomalies. The residual between the generated sample and the input is then used as the anomaly measure. Methods such as AnoGAN \cite{anogan}, EGBAD \cite{egbad}, and fast AnoGAN \cite{fanogan} have been developed based on this idea. GAN-based methods suffer from issues such as time inefficiency, failure to converge during training, and mode collapse \cite{mode}.

\blue{AE-based and GAN-based anomaly detection techniques use reconstruction error to measure anomalies. However, their training objective does not directly relate to the task of anomaly detection.} There are very few approaches that, similar to DSVDD discussed before, realize the AD task by directly minimizing an anomaly score, rather than sample reconstruction error, during the training phase.
Although DSVDD has shown superior performance to other algorithms on several AD tasks, it suffers from the hypersphere collapse problem. Examining DSVDD loss, shown below, can be beneficial for understanding this issue:
\begin{equation}
    min_W \frac{1}{n}\sum_{i=1}^n ||\phi(x_i,W)-c||^2 + \frac{\lambda}{2}\sum_{l=1}^L ||W^l||_F^2
    \label{eq:1}
\end{equation}
where $c$ is the center of the hypersphere, $\phi(.)$ denotes the output of the neural network, $W$ represents the network weights, and $\lambda$ is a hyperparameter that controls the contribution of the regularization term. As can be seen, the above equation has a trivial solution with $W=0$ and $c=0$. To avoid this solution, $c$ must be excluded from the set of optimization problem variables, i.e. it must be considered as a non-trainable hyperparameter and be set to a non-zero value prior to the training phase. In addition, the network biases must be set to zero; otherwise, all the data points will be mapped to the hypersphere center $c$. Furthermore, activation functions with non-zero upper or lower bounds cannot be employed. This is because a network unit with a bounded activation function can be saturated for all inputs having at least one feature with a common sign, thereby emulating a bias term in the subsequent layer, leading to a hypersphere collapse. All these constraints limit the performance of deep SVDD in many applications. \b{Interested readers can refer to \cite{ruff2018deep} for a more detailed explanation.}

\b{Few studies have attempted to address the hypersphere collapse in DSVDD by adding more regularizers or changing some parts of the model: Sendera et al. \cite{maziarka2020flowbased} proposed instantiating DSVDD with a flow-based model. Due to the nature of the flow-based model, the proposed idea can prevent the hypersphere collapse problem. On the other hand, since normalizing flows rely on invertible transformation, they would limit the expressive power of the model, which can lead to suboptimal performance, particularly on complex datasets \cite{Copula-Based}. In another work, Chong et al. \cite{collapse} recommended using two regularization terms for preventing hypersphere collapse: (i) adding random noise and (ii) defining a term to penalize small values of minibatch variance. They have shown that their method can prevent hypersphere collapse on several datasets. However, it is not evident if the same improvement can be achieved on other datasets. Furthermore, these models do not tune $c$ during training. Inspired by these findings, we propose an effective way to alleviate hypersphere collapse in DSVDD.}

\blue{Recently, self-supervised methods like CSI \cite{tack2020csi} and GEOM \cite{golan2018deep} have shown promising results in visual anomaly detection in terms of accuracy. However, besides the fact that they cannot be applied to other data types, such as time series, these methods require large batches and training epochs. As a result, they commonly need access to a large dataset and are very time-consuming to be trained. This can limit their application in many real-world problems. Furthermore, designing a good pretext task for self-supervision usually requires knowledge about datasets. For instance, rotation prediction can be a good pretext task for the CIFAR-10 dataset but not for MNIST because digits 6 and 9 can become equal with a rotation. Therefore, developing deep SVDD-based methods is still an active field of research.}

\section{Proposed Method}
\label{method}

Let $D$ be our dataset and $X \in D$ be the set of normal samples. Given a subset of $X_{train} \in X$, we would like to learn an anomaly scoring function $S(x): D \rightarrow \mathbb{R}$ such that a lower score denotes a higher probability of sample $x$ being in $X$, or equivalently, sample $x$ being normal.

As shown in Figure \ref{fig:overview}, the proposed DASVDD method consists of two major components: a deep autoencoder and a support-vector data descriptor (SVDD). The autoencoder first maps the input data to a lower-dimensional latent representation space and then attempts to reconstruct it. Simultaneously, the SVDD finds a data enclosing hypersphere with the minimum volume on the latent space of the AE. Therefore, the autoencoder learns to map the data into a hypersphere with minimum volume while still can reconstruct the original input. The anomaly score is calculated by combining the reconstruction error and the euclidean distance between the lower-dimensional latent representation of the sample from the center of the enclosing hypersphere.
The AE network and SVDD are trained using only the normal samples, i.e. $X_{train}$. Therefore, the bottleneck of the autoencoder learns a representation that models the underlying representation of the normal data.

\subsection{DASVDD Anomaly Score and Objective Function}
Let $h(.)$ and $g(.)$ be the encoding and decoding functions of the AE, respectively, and let $\theta_e$ and $\theta_d$ denote their corresponding parameters (weights and biases). Given an input sample $x$, the encoder calculates its latent representation $z = h(x;\theta_e)$.
Then the reconstructed output is $\hat x = g(z;\theta_d)$.

We define the anomaly score as a combination of reconstruction error and the distance of the latent representation from the center of the hypersphere. Formally, given a sample $x$, the anomaly score $S(x)$ can be written as:
\begin{multline}
\label{eq:2}
    S(x) =||\hat x-x||^2 + \gamma ||z-c^*||^2 \\= ||g(h(x;\theta_e^*);\theta_d^*) - x||^2 + \gamma ||h(x;\theta_e^*)-c^*||^2
\end{multline}
where $c$ denotes the center of the enclosing hypersphere, and $\gamma$ is a hyperparameter balancing contribution of the two terms. The star symbol `$^*$' denotes the optimum value of the corresponding parameter that is obtained after completing the algorithm training phase. From now on, we refer to the first term of equation (\ref{eq:2}) as the reconstruction error term and to the second term as SVDD term.

The objective of the network is to minimize the anomaly score on the normal data. Therefore, we can set the objective function equal to the anomaly score and define it as shown below in (\ref{eq:3}) where $n$ denotes the batch size.
\begin{equation}
\label{eq:3}
    \min_{\theta_e,\theta_d,c} \frac{1}{n}\sum_{i=1}^n ||g(h(x_i;\theta_e);\theta_d) - x||^2 + \gamma ||h(x_i;\theta_e)-c||^2
\end{equation}
\blue{In the above loss function, we can add a term $\lambda \|\Theta\|_F$ to include the weight decay regularization as well. Here, $\lambda$ is the weight decay hyperparameter and $\Theta$ denotes the matrix of the concatenation of the encoder and decoder's weights.}
Including the reconstruction error, i.e. the first term, in the objective function is to avoid converging to a trivial latent representation that the decoder cannot reconstruct the normal samples from.
The second term corresponds to the SVDD term, which is to penalize the radius of the hypersphere. By minimizing the second term, we, in fact, minimize the average distance of the sample from the hypersphere centre $c$, hence minimizing the volume of the hypersphere enclosing the normal data points.

Since the objective function of DASVDD has a reconstruction error term, the all-zero weights solution does not reduce the value of the objective function to the minimum possible error value, i.e. zero; therefore, no hypersphere collapse happens. Obviously, $\gamma$ should not be set to a very high value to ensure the effective contribution of the reconstruction error term, hence avoiding the all-zero weights and zero hypersphere center solution. A simple heuristic approach for choosing an appropriate $\gamma$ value is presented in Section \ref{gamma}.
Note that, unlike the DSVDD algorithm, the proposed model treats the center of the hypersphere $c$ as a trainable parameter that can be trained during training. 

By mapping the data into a lower-dimensional representation close to the center $c$, the network learns to extract the common factors of variation from normal data. Since anomalies have intrinsic differences, we expect our network to be unable to reconstruct them correctly and/or map them further from the hypersphere center.

\subsection{Optimization and Training}
One of the challenges of jointly training the network parameters with hypersphere center $c$ is that they are from different numerical ranges. As a result, using a single optimizer for both sets of parameters might be inefficient, and can lead to the loss function collapsing to either of the two loss terms.

To jointly train the AE and SVDD, we propose the following strategy: at each training epoch, first, \ use $\kappa$ ($0<\kappa<1$) portion of each batch to train the network parameters $\theta_e$ and $\theta_d$ while the hypersphere center $c$ is fixed; then, use the remaining samples to train the hypersphere center $c$ while keeping the network parameters fixed. This training procedure is summarized in Algorithm \ref{training}.
	\begin{algorithm*}[t]
		\caption{Training procedure of DASVDD}\label{training}
		\textbf{Input:} $X_{train}$, $c^{(0)}$, $\theta^{(0)}=\{\theta_e^{(0)},\theta_d^{(0)}\}$, $max\_epochs$, $n $, $\gamma$\\
	\textbf{Output: }  $c^*$ and $\theta^*=\{\theta_e^*,\theta_d^*\}$
		\begin{algorithmic}[1]
			\State Initialize Network Parameters $\theta = \{\theta_e,\theta_d\}$ and hypersphere center $c$
			\For{$j<$ $max\_epochs$}
			\State Pick $\kappa$ portion of the samples batch
			\State Optimize $\theta^{(j+1)}$ to minimize $ \sum_{i=1}^{[\kappa n]}\|x_i-\hat x_i\|^{2} + \gamma \sum_{i=1}^{[\kappa n]}\|h(x_{i};\theta_e^{(j+1)})-c^{(j)}\|^{2}$
			\State Pick the remaining $(1- \kappa)$ portion of the samples
			\State Optimize $c^{(j+1)}$ to minimize $\sum_{i=[\kappa n]+1}^{n}\|h(x_{i};\theta_e^{(j+1)})-c^{(j+1)}\|^{2}$
			\EndFor
		\end{algorithmic}
	\end{algorithm*}
We suggest using stochastic gradient descent (SGD) or its variants, such as Adam \cite{adam}, as the optimizer for the autoencoder's parameters. For training the hypersphere center $c$, we recommend using an algorithm with an adaptive learning rate, such as AdaGrad \cite{adagrad}. This usually results in faster convergence of the training procedure since it allows the assignment of higher weights to tune $c$ in the first few epochs. Experimental evaluations in Section \ref{Converge} demonstrate that this strategy results in a stable training process.

Moreover, if we fix the network parameters prior to training $c$, we can easily prove that the optimal $c$ can be calculated by averaging over the latent representation of the batch samples:

\begin{equation}
\label{c:eq}
    c = \frac{1}{|B|} \sum_{i=1}^{|B|} h(x_i;\theta_e)
\end{equation}
where $|B|$ is the batch size in the above equation. This shows that jointly training $c$ with the network parameters does not add any complexity to the optimization process. To prove the equation \ref{c:eq}, we start by writing the loss function as:

$$\mathcal{L} = \mathcal{L}_{RE} + \mathcal{L}_{DSVDD}$$
if the batch size is equal to $|B|$, the total loss of a batch can be written as:

$$\Rightarrow \mathcal{L} = \sum_{i=1}^{|B|} (\norm{x_i-\hat{x_i}}^2 + \norm{h(x_i;\theta_e) - c}^2)$$
now, we take the derivative of $\mathcal{L}$, and set it equal to zero.

$$\frac{\partial \mathcal{L}}{\partial c} = \frac{\partial}{\partial c} \sum_{i=1}^{|B|} (\norm{x_i-\hat{x_i}}^2 + \norm{h(x_i;\theta_e) - c}^2) = 0$$
since we assume that the network parameters, i.e. $\theta_e, \theta_d$ are fixed during tuning $c$, the derivative of the $\mathcal{L}_{RE}$ will be zero:

$$\Rightarrow \frac{\partial}{\partial c} \sum_{i=1}^{|B|} (\norm{h(x_i;\theta_e) - c}^2) = 0$$

$$\Rightarrow \sum_{i=1}^{|B|} \frac{\partial}{\partial c} (\norm{h(x_i;\theta_e) - c}^2) = 0$$

$$\Rightarrow \sum_{i=1}^{|B|} -2(h(x_i;\theta_e) - c) = 0$$

$$\Rightarrow \sum_{i=1}^{|B|} h(x_i;\theta_e) - \sum_{i=1}^{|B|} c = 0$$

$$\Rightarrow \sum_{i=1}^{|B|} h(x_i;\theta_e) - |B|c = 0$$

$$\Rightarrow c = \frac{1}{|B|} \sum_{i=1}^{|B|} h(x_i;\theta_e)$$

\subsection{Choice of Hyperparameter $\gamma$}
\label{gamma}
DASVDD uses hyperparameter $\gamma$ for balancing the two terms of the anomaly score (which is also our objective function). Depending on the dataset characteristics, network architecture and initial parameter values, the reconstruction error and SVDD terms might have different numerical ranges.
Therefore, a reasonable choice for $\gamma$ is a value proportional to the ratio of these two terms as below:
\begin{equation}
\label{eq:4}
    \gamma = \frac{1}{N}\sum_{i=1}^N \frac{||\hat x_i - x_i||^2}{||z_i - c^*||^2} = \frac{1}{N}\sum_{i=1}^N \frac{||g(h(x_i;\theta_e^*),\theta_d^*) - x_i||^2}{||h(x_i;\theta_e^*) - c^*||^2}
\end{equation}
where $N$ is the total number of training samples. Because the DASVDD anomaly score is also the method objective function, we do not have access to the trained optimal parameters prior to the the training outset. 
As such, we suggest using initial values of the network parameters (i.e. $\theta_e$ and $\theta_d$) and initial value of the hypersphere center $c$ instead of their optimum values when computing $\gamma$ in (\ref{eq:4}). We suggest randomly initializing $\theta_e$ and $\theta_d$ and initializing $c$ to zero. Because of the random initialization of the network parameters, we suggest repeating this procedure $T$ times, with $T$ different random initial values (where $c$ is set to zero), and using the average value as the final $\gamma$ to be used in the algorithm training phase. This approach is mathematically described in (\ref{eq:5}), where the superscript $^{(0)}$ denotes the initial value of the corresponding parameter.
\begin{equation}
    \gamma = \frac{1}{T}\sum_{t=1}^T \frac{1}{N}\sum_{i=1}^N \frac{||g(h(x_i;\theta_e^{(0)}),\theta_d^{(0)}) - x_i||^2}{||h(x_i;\theta_e^{(0)})-c^{(0)}||^2}
    \label{eq:5}
\end{equation}

\section{Experiments}
\label{results}

In this section, we test the effectiveness of our proposed anomaly detection method on seven publicly available benchmark datasets and compare its performance against a variety of state-of-art AD algorithms.

\subsection{Datasets}
\label{dataset}
We employ three publicly available computer vision benchmark datasets, two datasets from Outlier Detection DataSets (ODDS) repository \cite{odds}, \blue{an intrusion detection dataset (AWID 3) \cite{awid3}, and a dataset containing acoustic recordings from four industrial equipment (MIMII) \cite{MIMII}}. Previous AD works, such as \cite{ruff2018deep,abatiand}, commonly use computer vision datasets as benchmarks mainly because these are usually high-dimensional and let us assess results visually.
The employed datasets are as follows.
\textbf{(1)} MNIST \cite{mnist} which consists of 70,000 28$\times$28 handwritten digits monochrome images. 
\textbf{(2)} CIFAR10 \cite{cifar} which consists of 60,000 color images of size $32\times 32$ from 10 different objects.
\textbf{(3)} Fashion MNIST (FMNIST) \cite{fmnist} which has 70,000 grey scale images of size $28\times 28$ from 10 fashion products.
\textbf{(4)} ODDS Speech \cite{speech} which contains 3,686 segments of English speech spoken with different accents where segments are represented by a 400-dimensional feature vector called I-vector. The majority of samples are from American English accents, which are labelled as normal, and the rest of the data, which is equal to 1.65\% of samples, are labelled as anomalies. 
\textbf{(5)} PIMA \cite{uci}, which is included in the ODDS repository and is a subset of the "Pima Indians diabetes dataset" of the UCI repository. It contains data from female patients of at least 21 years old. The dataset has 765 samples, of which 268 (35\%) are anomalies. Each sample has eight attributes. 
\blue{\textbf{(6)} AWID 3 \cite{awid3}, a network intrusion detection dataset which concentrates on WPA2-Enterprise and Protected Management Frames (PMF). It includes $254$ features, and $21$ assaults from various types.}
\blue{\textbf{(7)} MIMII \cite{MIMII} comprises actual acoustic samples that are utilized for identifying faulty industrial equipment. The dataset comprises 10-second audio segments originating from four machine types: Fans, Pumps, Slide-Rails, and Valves. The signals are captured at a 16 KHz sampling rate.}
The CIFAR10 and FMNIST datasets are available under the MIT licence, and the MNIST, AWID 3 and ODDS datasets under CC BY-SA 3.0, Creative Commons Attribution 4.0, and Affero GPL 3.0 licences, respectively.

These datasets are from different domains of application and can help us to gain a better understanding of the DASVDD performance under different circumstances.
The MNIST, CIFAR10, and Fashion MNIST datasets all have ten classes. The full list of CIFAR10 and FMNIST classes is included in Table \ref{classlist}. In each experiment, we pick one of the classes as normal and label the rest as anomalies. Using this approach, we create ten one-class classifiers for each dataset. This approach is common in image anomaly detection and was previously used by similar studies \cite{ruff2018deep,abatiand}. We use the original training and test data and only employ normal samples for training the models. No pre-processing step is performed on the MNIST and Fashion MNIST datasets. However, similar to \cite{ruff2018deep}, we pre-processed the CIFAR10 images with global contrast normalization \cite{GoodBengCour16}. The same pre-processing is employed for our comparison algorithms. The \blue{rest of the datasets} are already labelled as being normal or anomalous and can be directly used for AD.

\begin{table}[t]
\caption{List of Classes for CIFAR-10 and Fashion MNIST datasets.}
\label{classlist}
\begin{center}
\begin{tabular}{ccc}
    \toprule
    \textbf{Dataset}&\textbf{Class}&\textbf{Single Class Name}\\
    \midrule
    {\multirow{10}{*}{CIFAR-10}}&0&Airplane\\
    &1&Car\\
    &2&Bird\\
    &3&Cat\\
    &4&Deer\\
    &5&Dog\\
    &6&Frog\\
    &7&Horse\\
    &8&Ship\\
    &9&Truck\\
    \midrule
    {\multirow{10}{*}{Fashion MNIST}}&0&T\-Shirt\\
    &1&Trouser\\
    &2&Pullover\\
    &3&Dress\\
    &4&Coat\\
    &5&Sandal\\
    &6&Shirt\\
    &7&Sneaker\\
    &8&Bag\\
    &9&Ankle Boot\\
    \bottomrule
\end{tabular}
\end{center}
\end{table}

\subsection{Competing Methods}

We compare performance of DASVDD against three commonly-used traditional baselines, i.e. one class SVM (OCSVM) \cite{ocsvm}, kernel density estimation (KDE) \cite{kde}, and isolation forest (IF) \cite{IF}, as well as \blue{eight} state-of-the-art deep anomaly detection algorithms, i.e. deep autoencoder (AE) \cite{ae}, deep variational autoencoder (VAE) \cite{vae}, deep autoencoding Gaussian mixture model (DAGMM) \cite{dagmm}, AnoGAN \cite{anogan}, PixCNN \cite{pixcnn}, autoregressive novelty detector (AND) \cite{abatiand}, One-Class GAN (OCGAN) \cite{Perera_2019_CVPR}, and deep SVDD (DSVDD) \cite{ruff2018deep}.

\blue{OCSVM is a popular classic kernel-based anomaly detection algorithm. It is similar to SVM, but instead of learning a hypersphere that separates the classes, it tries to find a hypersphere that encompasses all training points.} In our experiments, we used it with its default kernel, i.e. radial basis function (RBF) with the corresponding hyperparameter $\nu=0.5$.
KDE is a classic yet effective anomaly detection method with promising results even in challenging anomaly detection problems. The bandwidth parameter of the Gaussian kernel is tuned using 5-fold cross-validation.
IF is a non-deep method. We set its hyperparameters, i.e. the number of trees and sub-sampling size, to respectively $100$ and $256$, as suggested in the original paper.
AE with mean squared error loss is used as one of our deep learning-based baselines. We chose the same network architecture as our proposed method's autoencoder. The reconstruction error is used as an anomaly score.
VAE is another autoencoder-based method that employs a variational autoencoder instead of a simple AE for data reconstruction.
DAGMM is a model that uses an autoencoder along with a Gaussian mixture model on the autoencoder latent representation. It has shown competitive results on several anomaly detection datasets. 
AnoGAN is a relatively recent anomaly detection method, and one of the first GAN-based algorithms for AD.
\blue{PixCNN \cite{pixcnn} is a generative algorithm that models the image pixel-by-pixel using autoregressive connections.} AND is a recent anomaly detection algorithm that jointly learns the autoencoder parameters along with an autoregressive model on its latent representation. \blue{OCGAN \cite{Perera_2019_CVPR} is another state-of-the-art method which uses adversarial training and GAN framework to constrain the latent representation of a denoising autoencoder to represent the normal class.}
\begin{figure*}[t]
  \centering
  \includegraphics[width=0.9\linewidth]{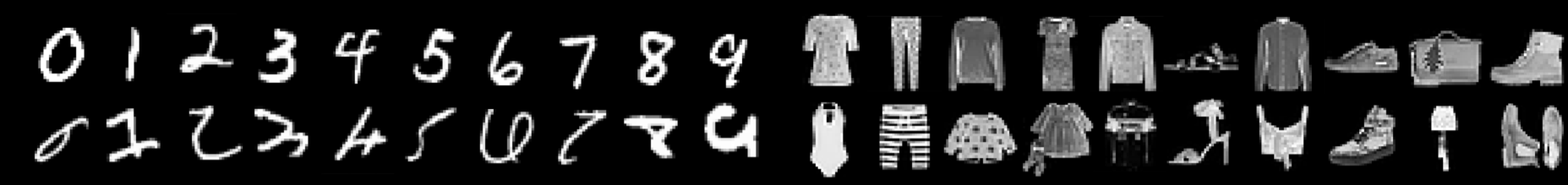}
  \caption{Samples with the lowest (top row) and highest (bottom row) anomaly scores in MNIST (10 left columns) and FMNIST (10 right columns).}
  \label{fig:fig2}
\end{figure*}

DSVDD is the closest algorithm to our proposed method. We use DSVDD as one of the baselines to assess if our modifications and the proposed method can yield better performance.
In terms of the compared deep methods, If the results of a method on some datasets are not reported, we run the released code with hyper-parameters mentioned in the original paper and report the results.

\blue{The goal of our approach is to detect anomalies regardless of their input type. Therefore, for the sake of fair comparison, we do not compare our algorithm to image-specific AD models\s{such as GT \cite{golan2018deep} and CSI \cite{tack2020csi}}. These models incorporate input-related modules, such as data augmentation, in their framework, making them unusable for anomaly detection on other data types \cite{golan2018deep,tack2020csi}.}

\begin{table*}[t]
\caption{AUC (\%) on benchmark datasets. Best performing method is denoted in boldface. The reported results are the average AUC over 10 runs. Standard deviations are reported for DASVDD}.
\label{resultstable}
\begin{center}
\begin{tabular}{c@{}c@{}cccccccccccc}
    \toprule
    Dataset & &OCSVM  &KDE &IF &AE &VAE &DAGMM &AnoGAN & \blue{PixCNN}& AND& \blue{OCGAN} & DSVDD & \textit{DASVDD}\\
    \midrule
    {\multirow{10}{*}{\normalsize{MNIST}}}&0&98.6&97.1&98.0&98.8&99.7&50.0&96.6&\blue{53.1}&98.4&\blue{99.8}&98.0&\textbf{99.7 $\pm$0.1}\\
    &1&99.5&98.9&97.3&99.3&99.9&76.6&99.2&\blue{99.5}&99.5&\blue{\textbf{99.9}}& 99.6&\textbf{99.9$\pm$0.0}\\
    &2&82.5&79.0&88.6&91.7&93.6&32.6&85.0&\blue{47.6}&94.7&\blue{94.2}&91.7&\textbf{95.4$\pm$0.6}\\
    &3&88.1&86.2&89.9&88.5&95.9&31.9&88.7&\blue{51.7}&95.2&\blue{\textbf{96.3}}&91.9&96.2$\pm$0.4\\
    &4&94.9&87.9&92.7&86.2&97.3&36.8&89.4&\blue{73.9}&96.0&\blue{97.5}&94.9&\textbf{98.1$\pm$0.1}\\
    &5&77.1&73.8&85.5&85.8&96.4&49.0&88.3&\blue{54.2}&97.1&\blue{\textbf{98.0}}&88.5&97.2$\pm$0.6\\
    &6&96.5&87.6&95.6&95.4&99.3&51.5&94.7&\blue{59.2}&99.1&\blue{99.1}&98.3&\textbf{99.6$\pm$0.1}\\
    &7&93.7&91.4&92.0&94.0&97.6&50.0&93.5&\blue{78.9}&97.0&\blue{\textbf{98.1}}&94.6&\textbf{98.1$\pm$0.3}\\
    &8&88.9&79.2&89.9&82.3&92.3&46.7&84.9&\blue{34.0}&92.2&\blue{93.9}&93.9&\textbf{94.2$\pm$1.2}\\
    &9&93.1&88.2&93.5&96.5&97.6&81.3&92.4&\blue{66.2}&97.9&\blue{98.1}&96.5&\textbf{98.3$\pm$0.5}\\
    \hdashline
    &\textit{avg:}&91.3&86.9&92.3&91.9&96.9&50.6&91.3&\blue{61.8}&96.7&\blue{97.5}&94.8&\textbf{97.7}\\
    \midrule
    {\multirow{10}{*}{\normalsize{CIFAR10}}}&0&61.6&61.2&60.1&59.9&62.0&41.4&67.1&\blue{\textbf{78.8}}&67.8&\blue{75.7}&61.7&68.6$\pm$0.7\\
    &1&63.8&64.0&50.8&63.8&\textbf{66.4}&57.1&54.7&\blue{42.8}&58.2&\blue{53.1}&65.9&64.3$\pm$0.6\\
    &2&50.0&50.1&49.2&50.9&38.2&53.8&52.9&\blue{61.7}&51.7&\blue{\textbf{64.0}}&50.8&55.8$\pm$0.8\\
    &3&55.9&56.4&55.1&59.4&58.6&51.2&54.5&\blue{57.4}&57.9&\blue{\textbf{62.0}}&59.1&58.6$\pm$0.2\\
    &4&66.0&66.2&49.8&59.8&38.6&52.2&65.1&\blue{51.1}&65.4&\blue{\textbf{72.3}}&60.9&64.0$\pm$0.2\\
    &5&62.4&62.4&58.5&63.2&58.6&49.3&60.3&\blue{57.1}&64.3&\blue{62.0}&\textbf{65.7}&62.6$\pm$0.5\\
    &6&74.7&74.9&42.9&65.2&56.5&64.9&58.5&\blue{42.2}&61.3&\blue{\textbf{72.3}}&67.7&71.0$\pm$0.1\\
    &7&62.6&62.6&55.1&65.1&62.2&55.3&62.5&\blue{45.4}&63.0&\blue{57.5}&\textbf{67.3}&64.6$\pm$0.2\\
    &8&74.9&75.1&74.2&76.9&66.3&51.9&75.8&\blue{71.5}&73.9&\blue{\textbf{82.0}}&75.9&81.1$\pm$0.4\\
    &9&75.9&\textbf{76.0}&58.9&72.7&73.7&54.2&66.5&\blue{42.6}&69.7&\blue{55.4}&73.1&73.7$\pm$0.3\\
    \hdashline
    &\textit{avg:}&64.8&64.9&55.5&63.7&58.1&53.1&61.8&\blue{55.1}&63.3&\blue{65.7}&64.8&\textbf{66.5}\\
    \midrule
    {\multirow{10}{*}{\normalsize{FMNIST}}}&0&90.6&88.3&86.8&71.6&87.4&51.9&89.0&\blue{78.9}&88.3&\blue{90.1}&79.1&\textbf{91.2$\pm$0.1}\\
    &1&97.5&94.3&97.7&96.9&97.7&34.0&97.1&\blue{90.2}&96.4&\blue{98.2}&94.0&\textbf{99.0$\pm$0.0}\\
    &2&88.1&87.7&87.1&72.9&81.6&26.9&86.5&\blue{80.8}&86.8&\textbf{\blue{89.8}}&83.0&89.3$\pm$0.1\\
    &3&91.3&88.4&90.1&78.5&91.2&57.0&91.2&\blue{84.5}&92.2&\blue{92.0}&82.9&\textbf{93.7$\pm$0.1}\\
    &4&88.5&86.3&89.8&82.9&87.2&50.4&87.6&\blue{83.2}&88.0&\blue{90.3}&87.0&\textbf{90.7$\pm$0.4}\\
    &5&87.6&85.9&88.7&93.1&91.6&70.5&89.6&\blue{81.6}&86.8&\blue{89.1}&80.3&\textbf{93.8$\pm$1.3}\\
    &6&81.4&74.7&79.7&66.7&73.8&48.3&74.3&\blue{72.1}&76.9&\blue{79.3}&74.9&\textbf{82.8$\pm$0.1}\\
    &7&98.4&96.1&98.0&95.4&97.6&83.5&97.2&\blue{92.9}&98.8&\textbf{\blue{98.8}}&94.2&98.6$\pm$0.1\\
    &8&86.0&84.6&88.3&70.0&79.5&55.1&81.9&\blue{79.4}&87.5&\blue{88.0}&79.1&\textbf{89.4$\pm$0.4}\\
    &9&97.7&94.2&97.9&80.7&96.5&34.0&89.9&\blue{84.2}&96.8&\blue{96.4}&93.2&\textbf{97.9$\pm$0.1}\\
    \hdashline
    &\textit{avg:}&90.7&88.0&90.6&80.9&88.4&51.8&88.4&\blue{82.8}&89.8&\blue{91.2}&84.8&\textbf{92.6}\\
    \midrule
    \normalsize{Speech}&&49.2&52.2&50.0&61.3&62.1&59.3&56.2&\blue{52.5}&57.4&\blue{53.9}&58.3&\textbf{62.4$\pm$1.1}\\
    \midrule
    \normalsize{PIMA}&&56.0&64.7&65.0&57.0&61.0&67.3&62.4&\blue{58.2}&64.5&\blue{66.6}&55.2&\textbf{72.2$\pm$1.2}\\
    \midrule
    \normalsize{\blue{AWID 3}}&&\blue{95.2}&\blue{94.3}&\blue{95.8}&\blue{95.3}&\blue{96.0}&\blue{93.1}&\blue{93.6}&\blue{92.3}&\blue{95.3}&\blue{96.0}&\blue{95.9}&\textbf{\blue{96.3$\pm$0.7}}\\
    \midrule
    \normalsize{\blue{MIMII}}&&\blue{56.8}&\blue{60.8}&\blue{59.2}&\blue{61.3}&\blue{62.8}&\blue{57.4}&\blue{60.9}&\blue{57.2}&\blue{63.5}&\blue{61.4}&\blue{64.2}&\textbf{\blue{67.8$\pm$2.3}}\\
    \bottomrule
\end{tabular}
\end{center}
\end{table*}

\subsection{DASVDD Implementation Details} 

\label{implement}
For all datasets, we use an autoencoder with fully connected layers. For both MNIST and FMNIST, we use one hidden layer of size 1,024. For CIFAR10 and Speech datasets, we used an encoder with two hidden layers of sizes 1,024 and 512, respectively. For all datasets except PIMA, the latent space size is set to 256.
Since the PIMA dataset has only a few attributes, we use a model with two hidden layers of size ten and set the latent size to 4.
In all datasets, we use leaky ReLU as the activation function and set the training batch size to 200. For optimizing the network parameters, we use Adam optimizer with an initial learning rate of 0.001. For optimizing $c$, we employ AdaGrad \cite{adagrad} optimizer with an initial learning rate of 1 and decay of 0.1. The center of hypersphere $c$ is randomly initialized from a normal distribution.
\b{In all our experiments, we set $\kappa = 0.9$}, the weight decay hyperparameter $\lambda = 10^{-7}$, and run the training for $300$ epochs. 
The $\gamma$ value for every dataset is calculated, prior to training, as is discussed in Section \ref{gamma} where $T$ is set to 10. The obtained $\gamma$ , see (\ref{eq:5}), for datasets MNIST, CIFAR10, FMNIST, Speech, PIMA, AWID 3, and MIMII are respectively 0.75, 5.22, 0.28, 2.56 and 713.64, 12.42, 0.92.
All algorithms were implemented in Python using the PyTorch framework. \b{All codes are run on Google Colaboratory GPU (Tesla K80) with 12GB RAM, and NVIDIA RTX A5000.}

\begin{figure*}[t]

\begin{minipage}[b]{.48\linewidth}
  \centering
  \centerline{\includegraphics[width=7.0cm]{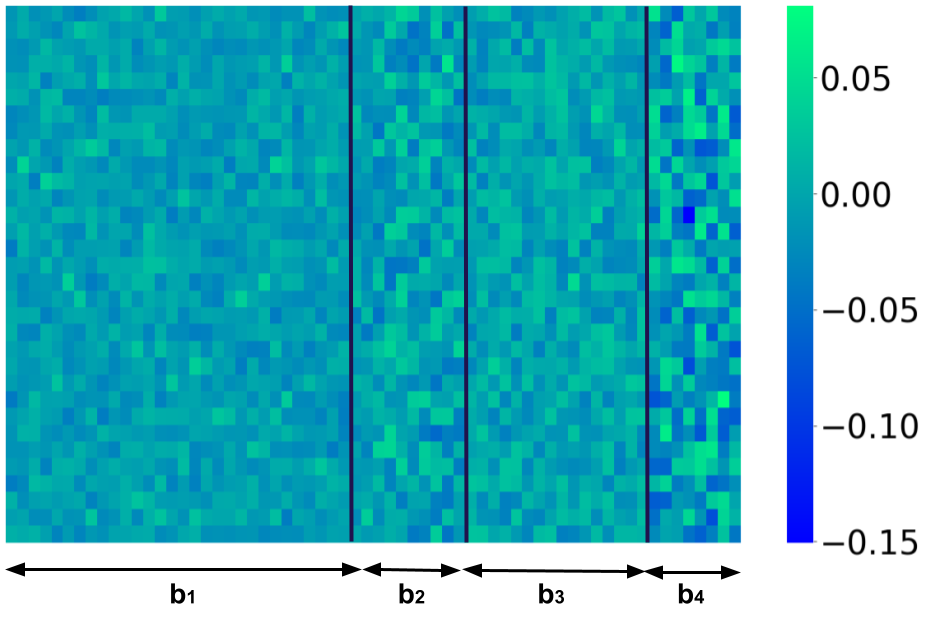}}
  \centerline{(a) Biases}\medskip
\end{minipage}
\begin{minipage}[b]{0.48\linewidth}
  \centering
  \centerline{\includegraphics[width=7.0cm]{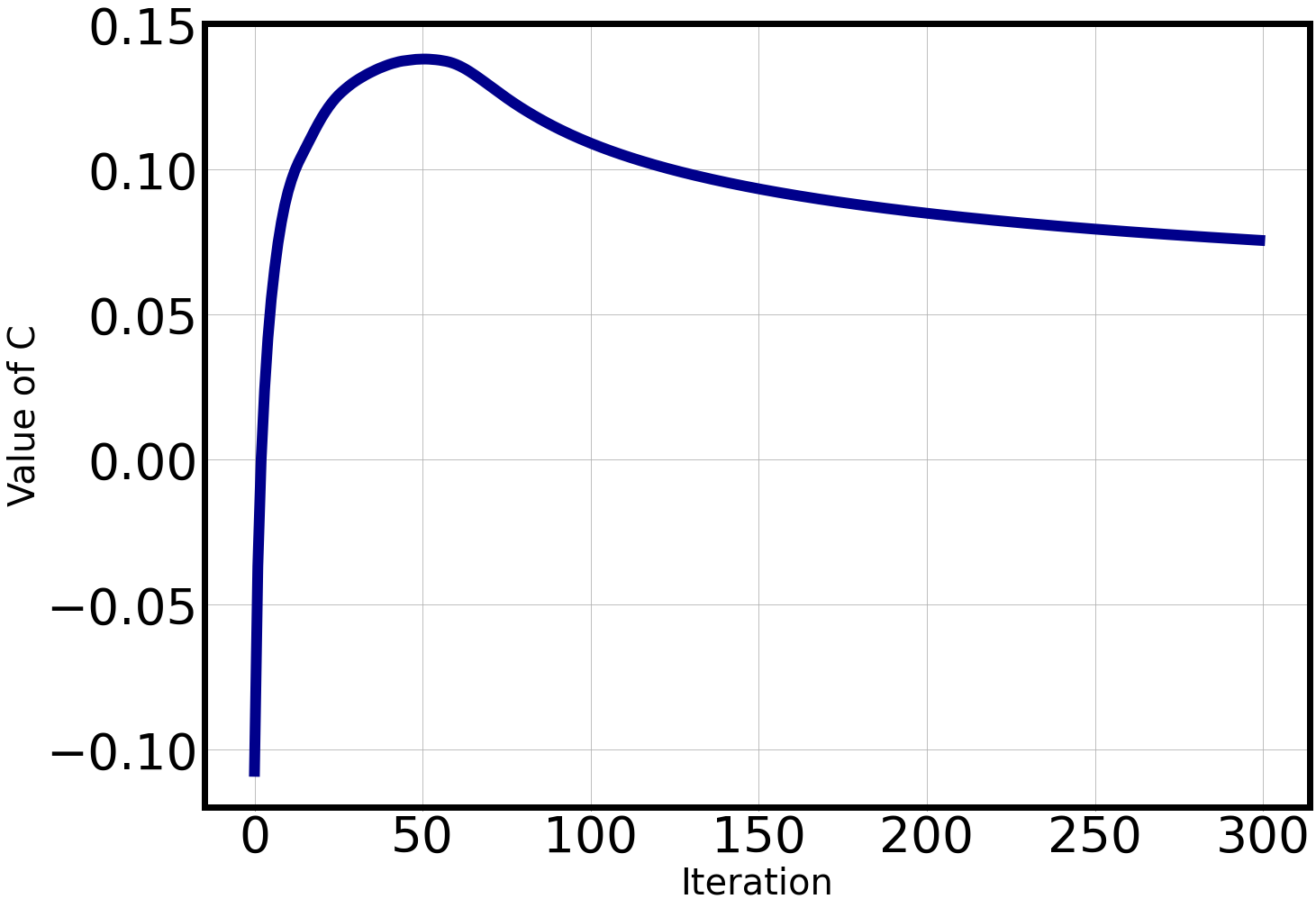}}
  \centerline{(b) Hypersphere Center}\medskip
\end{minipage}

\caption{(a) Visualization of the DASVDD encoder's weights where biases vector are reshaped and concatenated together for better visualization. Training data is class 8 of CIFAR-10. (b) Hypersphere Center versus iterations.}
\label{fig:biasc}
\end{figure*}

\subsubsection{Evaluation Metric} 
 A widely used metric for performance evaluation in the AD task is the area under receiver operator curve (ROC) \cite{ruff2018deep,abatiand}, which is known as AUC. We use AUC to measure the performance of our model and compare it with other AD methods. To achieve more reliable results, we repeat each experiment for ten runs and report the average and standard deviation of the calculated AUCs.

\subsection{Results and Analysis}

Results of our method, along with the baselines, are shown in Table \ref{resultstable}. In each case, the result of the best-performing algorithm is denoted in boldface.
Several observations can be made from this table:
\textbf{(1)} The average performance of the proposed DASVDD method is better than the baselines in all \blue{seven} datasets.
\textbf{(2)} In most classes of MNIST and FMINST datasets, all baselines except DAGMM can achieve a high average accuracy. Even in these cases, DASVDD has a slight edge over all baselines. On average, the performance of our model, along with other baselines, is better on MNIST rather than FMINST. This can stem from the fact that the FMINST is slightly more challenging due to the considerable amount of intra-class variances. If we inspect the results of class 1, which corresponds to `Trouser,' we can see that DASVDD significantly performs better than the baselines. Interestingly, this class is one of the only classes that have no visual similarity to any of the other nine classes. Other classes, such as `Sandal' and `Sneaker' or `Shirt' and `T-shirt' have similar samples, making it difficult to identify some near out-of-class instances as anomalies.
\textbf{(3)} \b{In Fashion MNIST and MNIST, the images are greyscaled, and their background is also removed. However, in the CIFAR-10, the diversity of the classes is higher, and the object of the class is presented in a complex background. Therefore, it is more challenging for the models, including ours, to extract meaningful patterns from the data in an unsupervised fashion. Since our model is not specifically tailored for learning geometrical features from images, the expressive power of the network will limit the model's performance on CIFAR-10. A user might apply augmentations to gain better performance on CIFAR-10 since it helps the network to capture more variations of the normal class. In general, it has been shown to improve the anomaly detection accuracy on this dataset \cite{golan2018deep}, but it is out of the scope of this paper since our model and other baselines are not concerned with data augmentation.}
\s{REWRITE, BETTER EXPLAIN: all models have a poor performance on CIFAR10. Yet, DASVDD beats other baselines in terms of average performance. It also performs better in several individual classes. CIFAR10 is considerably more challenging because of its complexity. Unlike FMINST and MNIST, CIFAR10 images are not well aligned, and they usually contain other objects in their background. Therefore, they are more complex and challenging. This can explain why all models have poorer performance on this dataset.}
\textbf{(4)} \blue{On the AWID 3 dataset, our model can outperform other baselines and achieve high accuracy. The same trend can also be observed in the MIMII dataset. These results show that our model can detect anomalies in non-image datasets as well}. On PIMA and Speech datasets of the ODDS repository, DASVDD still performs better than baselines. On the speech dataset, some baselines, such as OCSVM and IF, cannot perform better than a chance-level classifier. \b{In general, the performance of all models in this dataset, including ours, is lower than the performance in computer vision datasets. Given the inherent difficulty of the task, the results are justifiable.}
\textbf{(5)} Overall, DASVDD shows a robust performance across all datasets and classes. Some models, such as DAGMM, completely fail in several cases, such as in image datasets, particularly because they are not designed for these types of data. However, DASVDD can achieve an acceptable performance compared to other baselines in all tasks regardless of the data type or complexity of the problem.
\textbf{(6)} Figure \ref{fig:fig2} shows 
the most normal samples (i.e. samples with the lowest anomaly score) and the most anomalous samples (i.e. those with the most anomaly score) of the test set of the MNIST and FMNIST obtained using the proposed DASVDD method. We can see that the samples that our model detects as most anomalous are also deceptive and hard to detect for even human eyes.

\begin{figure*}[t]

\begin{minipage}[b]{.3\linewidth}
  \centering
  \centerline{\includegraphics[width=5.0cm]{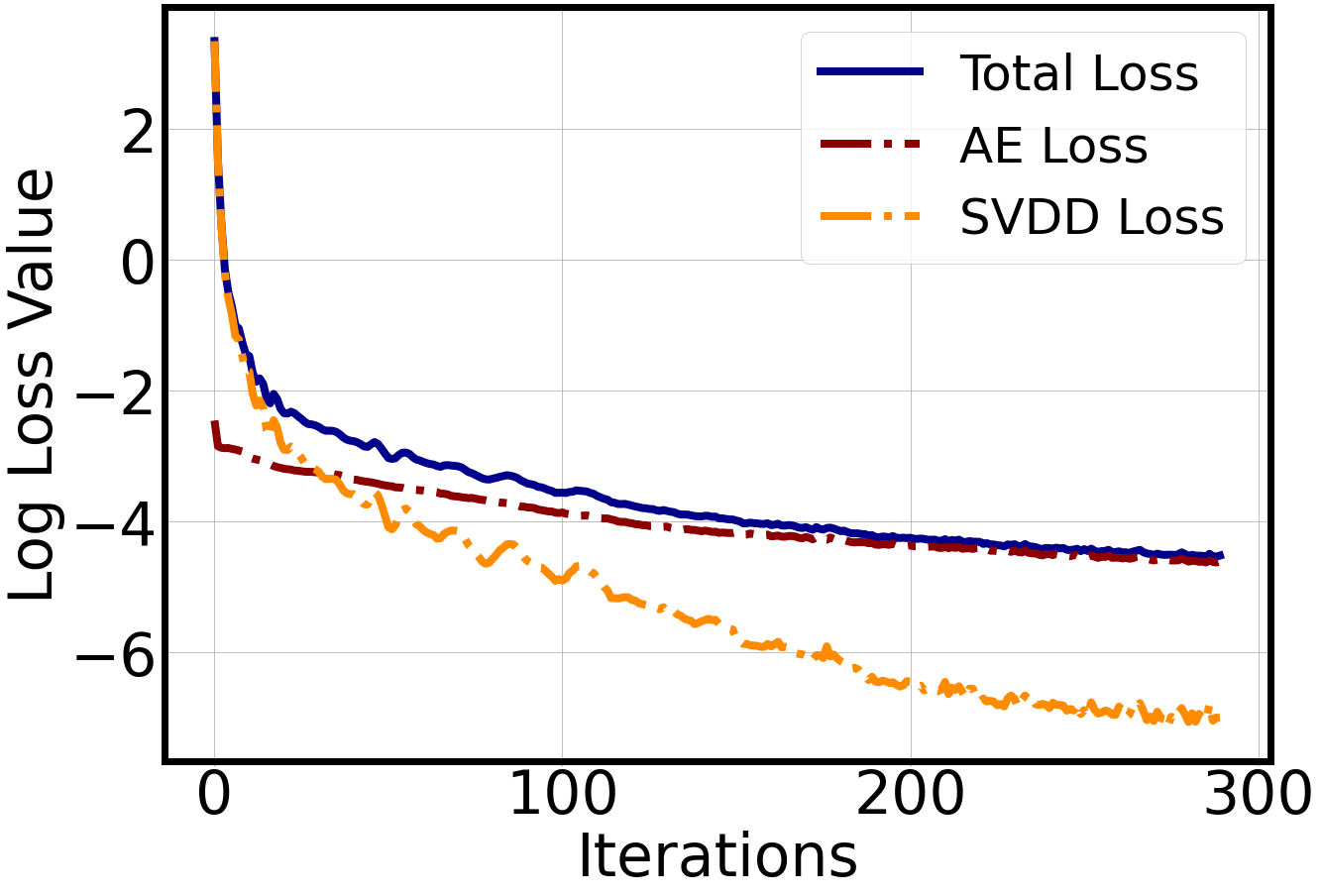}}
  \centerline{(a) MNIST}\medskip
\end{minipage}
\begin{minipage}[b]{0.3\linewidth}
  \centering
  \centerline{\includegraphics[width=5.0cm]{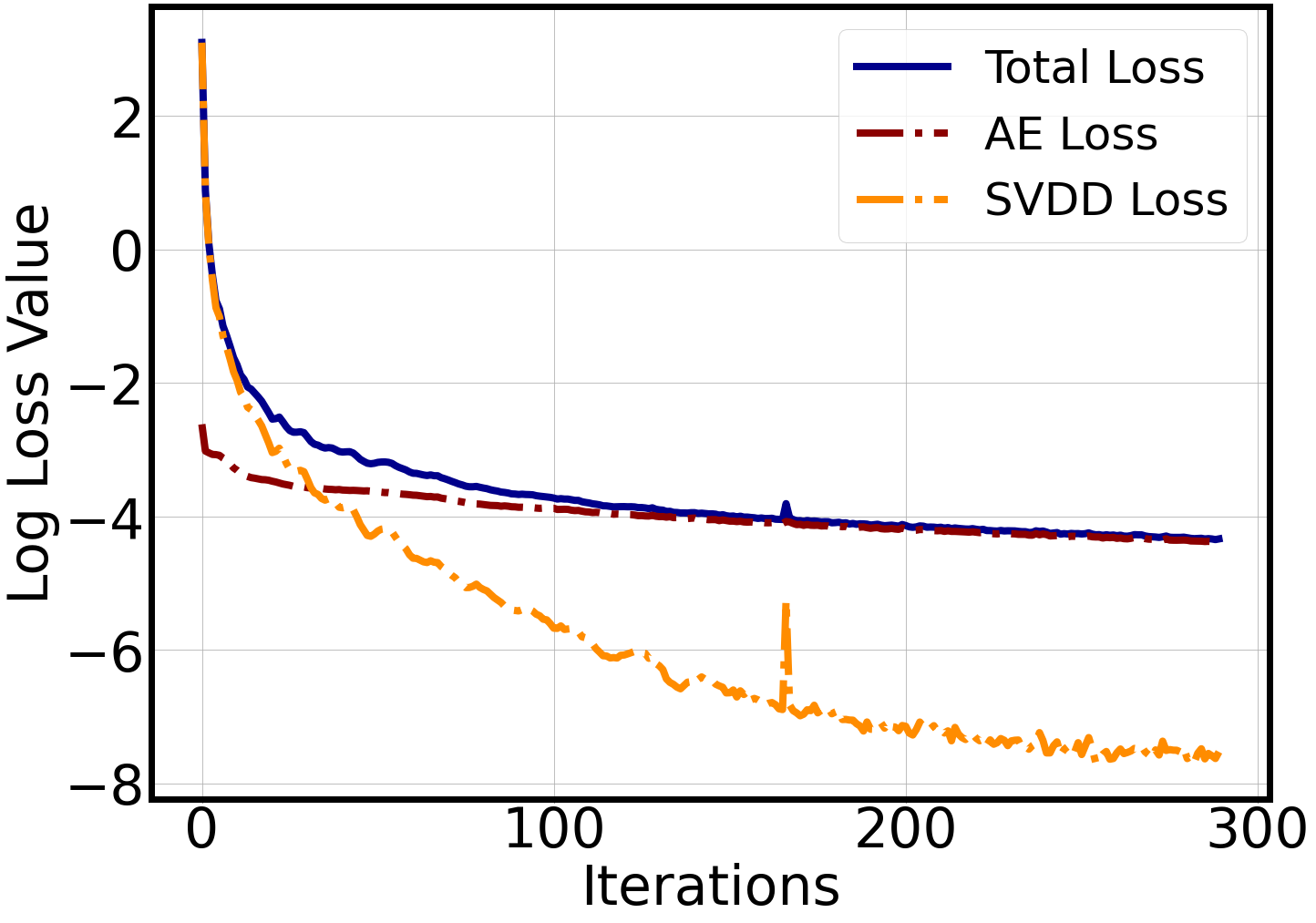}}
  \centerline{(b) Fashion MNIST}\medskip
\end{minipage}
\begin{minipage}[b]{0.3\linewidth}
  \centering
  \centerline{\includegraphics[width=5cm]{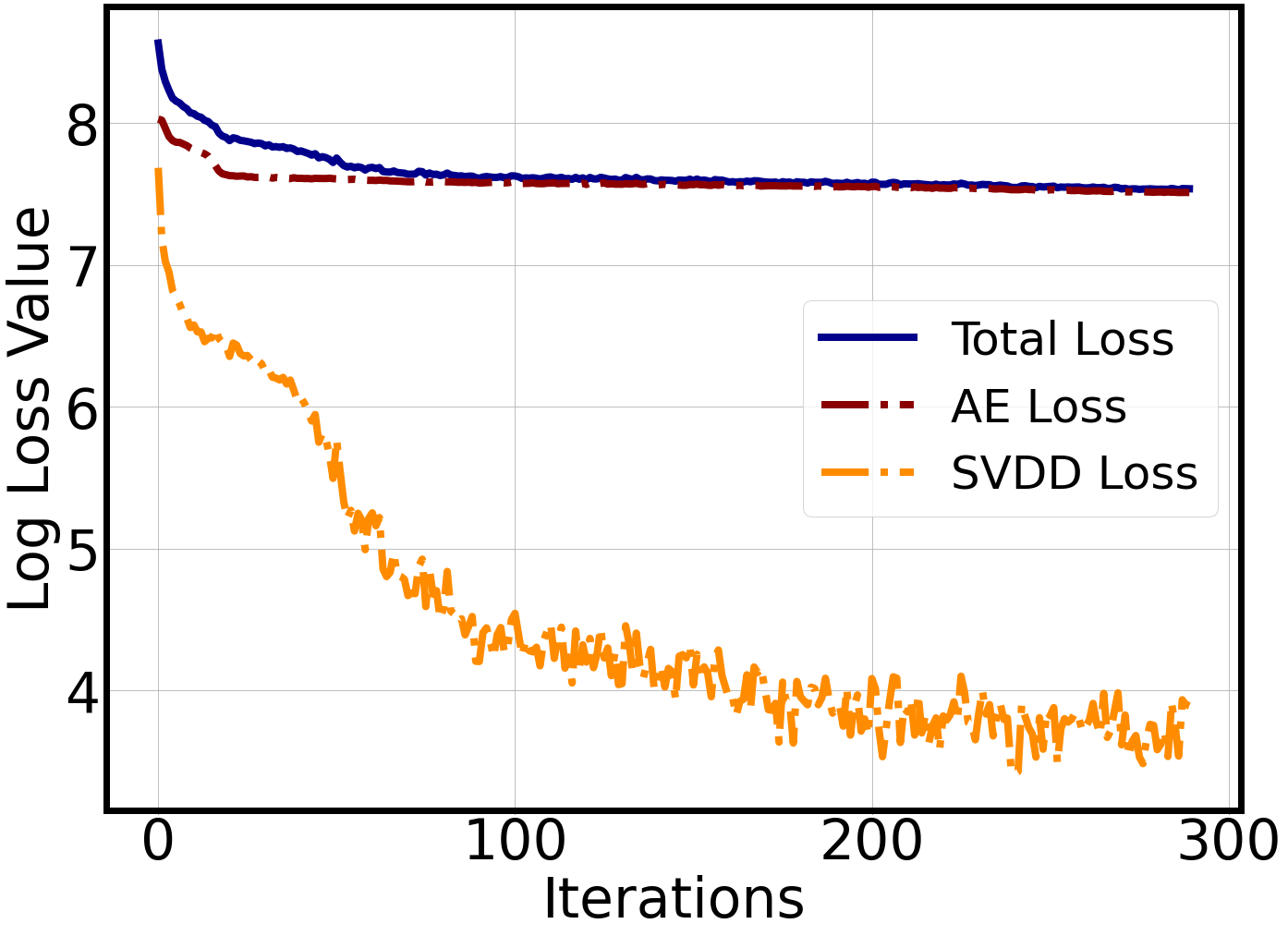}}
  \centerline{(c) CIFAR-10}\medskip
\end{minipage}

\caption{Total Loss, Reconstruction (AE) Loss, and Deep SVDD Loss for the class 8 of (a) MNIST, (b) Fashion MNIST, and (c) CIFAR-10 Datasets}
\label{fig:loss}
\end{figure*}

\subsection{Analysis of Biases and Hypersphere Center}

As discussed before, despite the conventional DSVDD method, the proposed DASVDD method does not run into the hypersphere collapse issue, and the hypersphere centre and the network biases are trainable parameters. We plot the value of $c$ versus the iteration and the values of biases during training where the training data is class 8 of CIFAR-10. Figure \ref{fig:biasc} shows the resulting plot. We can readily confirm that $c$ is getting trained and converges to its final value after almost $150$ iterations. Figure \ref{fig:biasc} (a) shows that the biases are non-zero. Yet, our model converges to a non-trivial solution (i.e., no hypersphere collapse is happening). This shows that without fixing $c$ and without setting biases to zero, our end-to-end trained network converges to a proper solution, which is the great advantage of DASVDD over all other DSVDD-based methods.

\blue{To further assess the impact of a trainable hypersphere center on model performance, we conducted additional experiments using a fixed parameter $c$ throughout the training process. We followed the approach of DSVDD \cite{ruff2018deep}, which involved performing a single forward pass and setting $c$ equal to the network's output mean. However, unlike our proposed method, $c$ remained constant during training. Table \ref{tab:c} presents the results of the model with a fixed $c$ alongside the original DASVDD results. Our findings indicate that fixing the hypersphere center resulted in lower performance, which was not surprising given that it restricted the subspace of possible optimal solutions.}

\begin{figure}[t]
  \centering
  \includegraphics[width=0.8\linewidth]{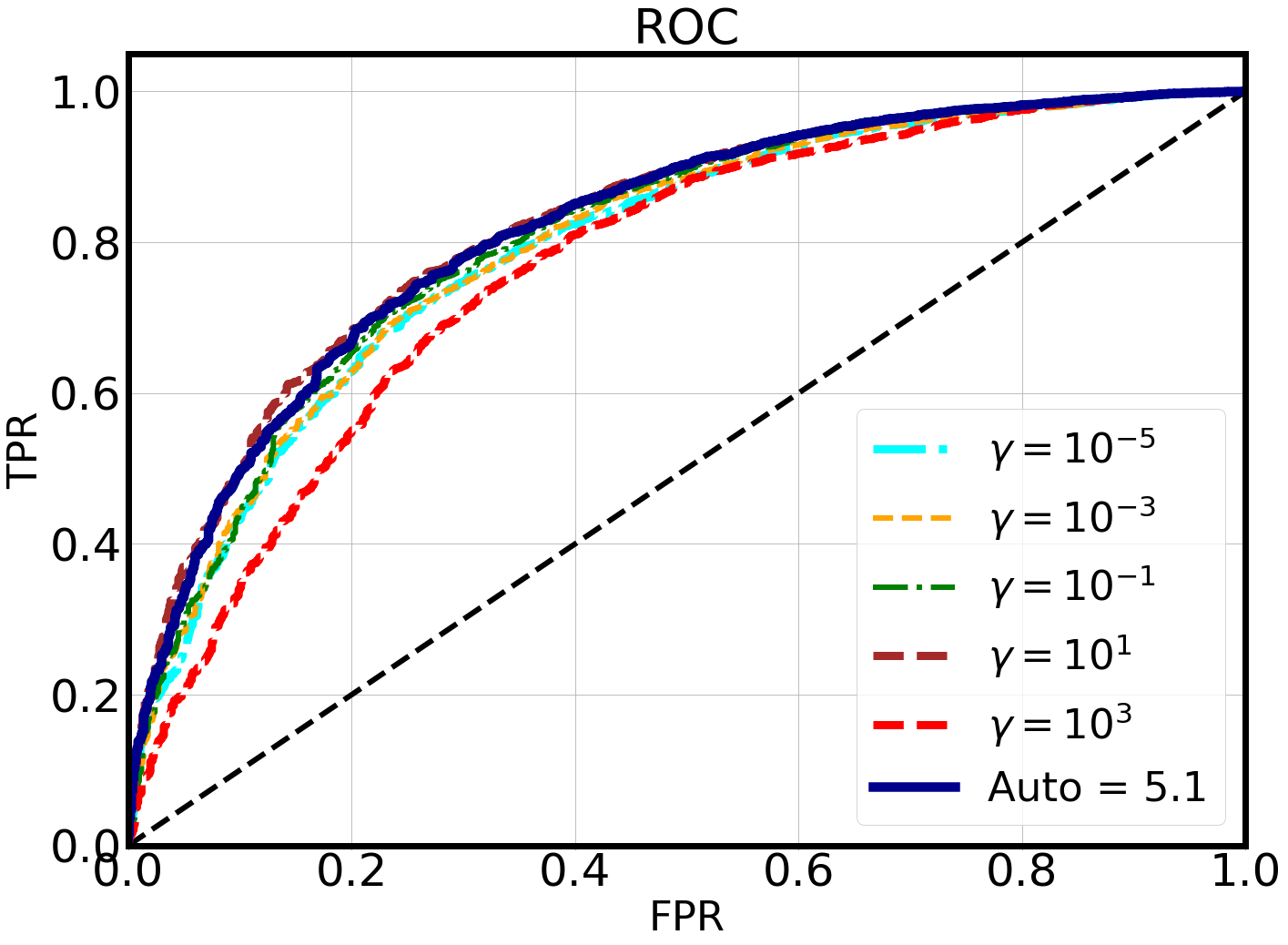}
  \caption{Receiver Operator Curve (ROC) for different values of $\gamma$ in class 8 of the CIFAR-10 dataset. The area under the curve (AUC) is an indicator of the performance of each model. The $\gamma$ value which we obtained using the approach described in Section \ref{gamma} is denoted by the thick blue curve.}
  \label{fig:gamma}
\end{figure}

\begin{table}[t]
\renewcommand{\arraystretch}{1.5}
\begin{center}
\begin{tabular}{ c|c|c|c } 
 \hline
 & MNIST & CIFAR10 & FMNIST\\
 \hdashline
 DASVDD (Trainable $c$) & $97.7\%$ & $66.5 \%$& $92.6 \%$ \\ 
 DASVDD (Fixed $c$) & $95.2\%$ & $64.7\%$ & $89.7 \%$ \\ 
 \hline
\end{tabular}
\end{center}
\label{tab:c}
\caption{\blue{Performance of the model with fixed and trainable $c$ on different datasets.}}
\end{table}

\subsection{Effect of Hyperparameters $\gamma$ and $T$}
In this section, we investigate the effect of hyperparameter $\gamma$ and demonstrate the effectiveness of the proposed $\gamma$ selection strategy discussed in Section \ref{gamma}. To this end, in Figure \ref{fig:gamma}, we plotted the ROC curve of our proposed anomaly detection method for $\gamma \in \{10^{-5},10^{-3},10^{-1},10,10^{3}\}$ along with the ROC curve obtained with our proposed automatic selection strategy shown in (\ref{eq:5}), where class 8 of CIFAR10 is used as the normal class.
We can observe that the choice of $\gamma$ can affect the performance of our anomaly detection model \blue{on the two extreme ends}. For large values of $\gamma$, such as $\gamma=10^{3}$,  which could be considered as a case equivalent to removing the effect of the reconstruction error loss in (\ref{eq:3}), the area under the curve (AUC) decreases significantly. Also, we can confirm that the value of $\gamma$ that we calculated using our proposed $\gamma$ selection method (shown as Auto) yields a ROC curve with at least a similar performance compared to other values. This observation, combined with the results of Table \ref{resultstable}, shows that our strategy for automatic selection of $\gamma$ yields an acceptable performance. Though the proposed $\gamma$ selection strategy is a great way to avoid the tedious manual tuning of the method's hyperparameter, we cannot claim that it finds the most optimal value for $\gamma$ in every task.
\b{Still, for a wide range of $\gamma$ values, in this example, from $\gamma = 10^{-3}$ to $\gamma = 10$, we can observe that the performance is robust and is not significantly affected. This shows that our model is not hypersensitive to this hyperparameter, which could explain our empirical observation of the success of the proposed heuristic in all datasets.} Finding a more efficient method for choosing $\gamma$ can be subject to further studies.

\begin{figure}[t]
  \centering
  \includegraphics[width=0.8\linewidth]{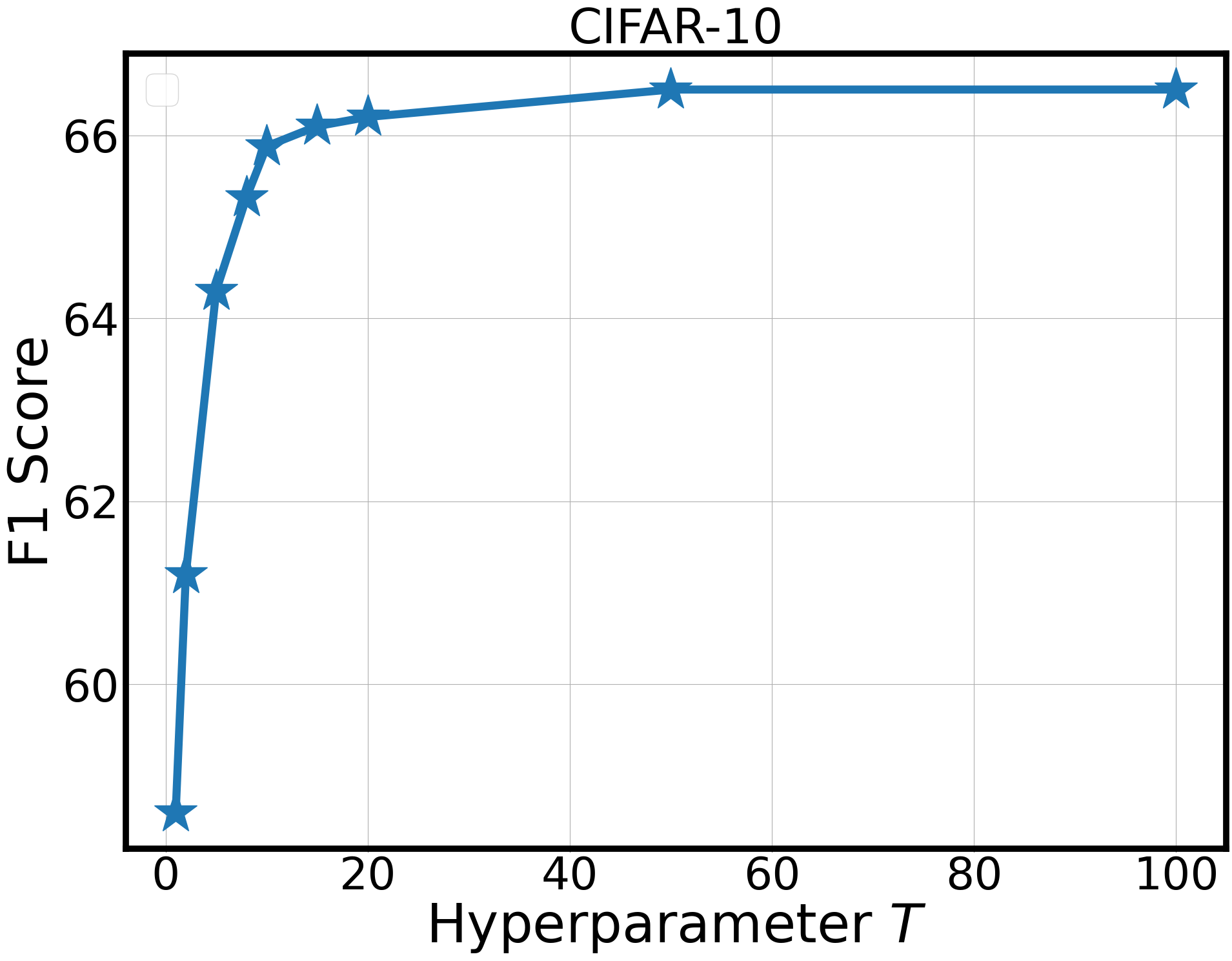}
  \caption{\blue{Performance of the model on the CIFAR-10 dataset for different values of $T$.}}
  \label{fig:tval}
\end{figure}

\blue{Our proposed heuristic approach introduces an additional hyperparameter $T$, which is the number of runs that we perform to find $\gamma$. Throughout our experiments, we fixed the value to $T=10$. To evaluate the sensitivity of our model, we ran a series of experiments on different classes of the CIFAR-10 dataset and changed the value of $T$. Figure \ref{fig:tval} shows the resulting graph. We can confirm that the performance improves by increasing $T$. However, for values higher than approximately $T=10$, the slope of the figure significantly decreases. Since increasing the value of $T$ increases the time needed to run the code, choosing a value on the elbow of this curve is a reasonable choice.}

\subsection{Convergence Analysis}
\label{Converge}
To gain better insight into the behaviour of our model, we plotted the value of the total loss, i.e. the objective function of \eqref{eq:3}, along with the value of its sub-components, i.e. the SVDD and reconstruction losses, versus iteration during training in Figure \ref{fig:loss}. We plotted the figure for class 8 of each dataset.
From Figure \ref{fig:loss}, we can confirm that the total loss converges during training. In addition, in all three datasets, the value of the total loss is close to the SVDD loss in the first few iterations and after that, it becomes almost similar to the reconstruction loss. This stems from the way we train our model. We use an optimizer with a large initial learning rate and a large decay parameter for training the hypersphere center $c$ during training. As a result, in the first few iterations, the SVDD loss abruptly decreases, but then, due to the learning rate decay, it changes more slowly. This is also reflected in the total loss. After the first few iterations, the AE-related loss term, i.e. the reconstruction error, dominates the loss function.

\subsection{\blue{Effect of Hyperparameter $\kappa$}}

\begin{figure}[t]
  \centering
  \includegraphics[width=0.8\linewidth]{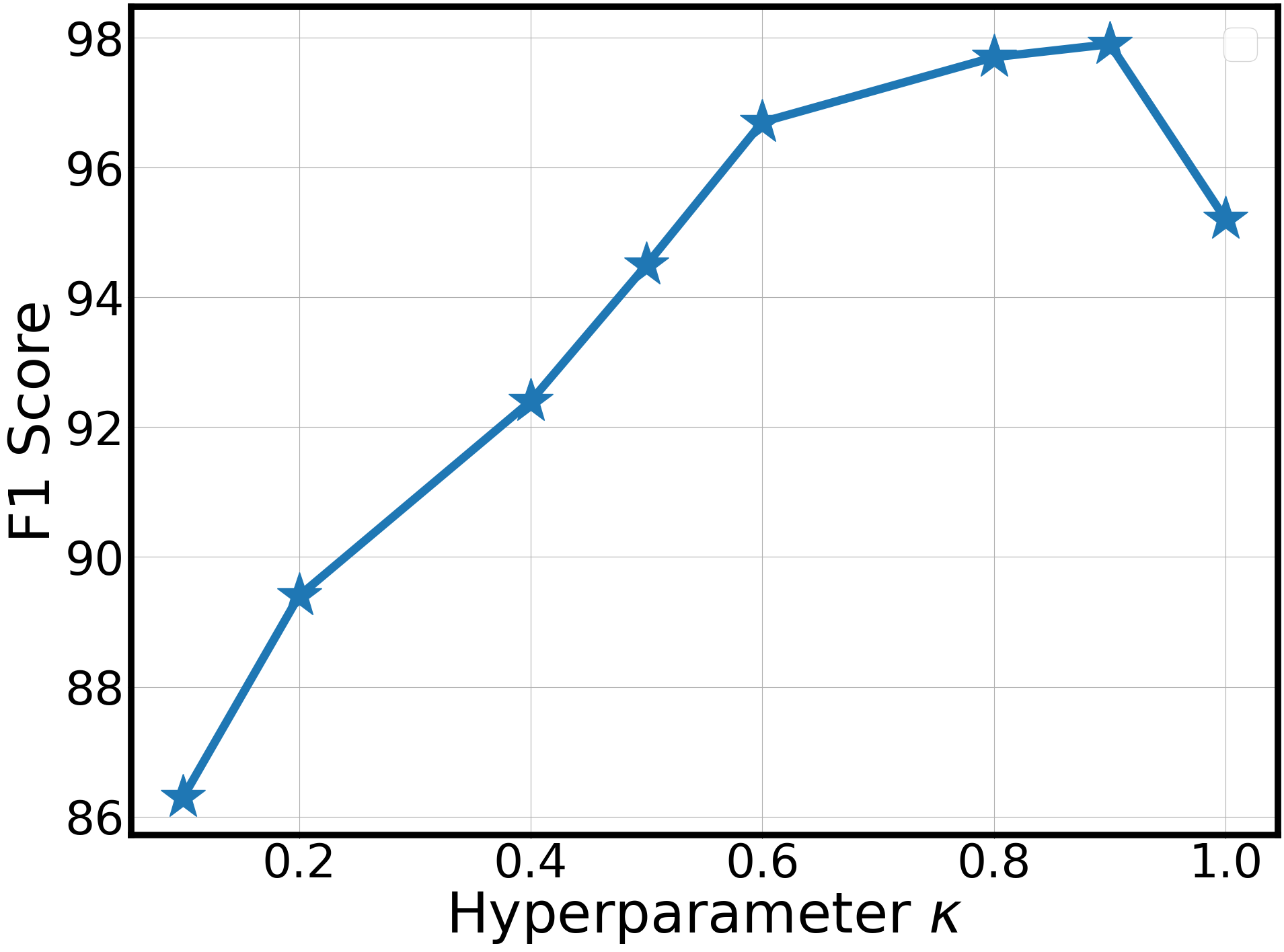}
  \caption{\blue{Effect of $\kappa$ on the model's performance on MNIST Dataset.}}
  \label{fig:effectofk}
\end{figure}

\blue{The hyperparameter $\kappa$ balances the portion of data which is being used for training the autoencoder and the hypersphere center. Throughout our paper, we set $\kappa = 0.9$ and empirically showed that it will yield a good performance across different datasets. In this section, we study the sensitivity of our model to this hyperparameter.}

\blue{To this end, we train and test the model on the MNIST dataset for different values of $\kappa$. Figure \ref{fig:effectofk} shows the performance versus the value of $\kappa$. In the MNIST dataset, the default value of $\kappa = 0.9$ yields the best performance. If we set $\kappa = 1$, we effectively fix the value of $c$, which downgrades the model performance. On the other hand, setting $\kappa$ to a small value means that fewer samples are available for training the autoencoder, which can also downgrade the model's performance significantly. Setting $\kappa = 0.9$ ensures that we have enough samples to train the autoencoder in each epoch.}

\subsection{Effect of the Latent Representation Size}

\begin{figure}[t]
  \centering
  \includegraphics[width=0.8\linewidth]{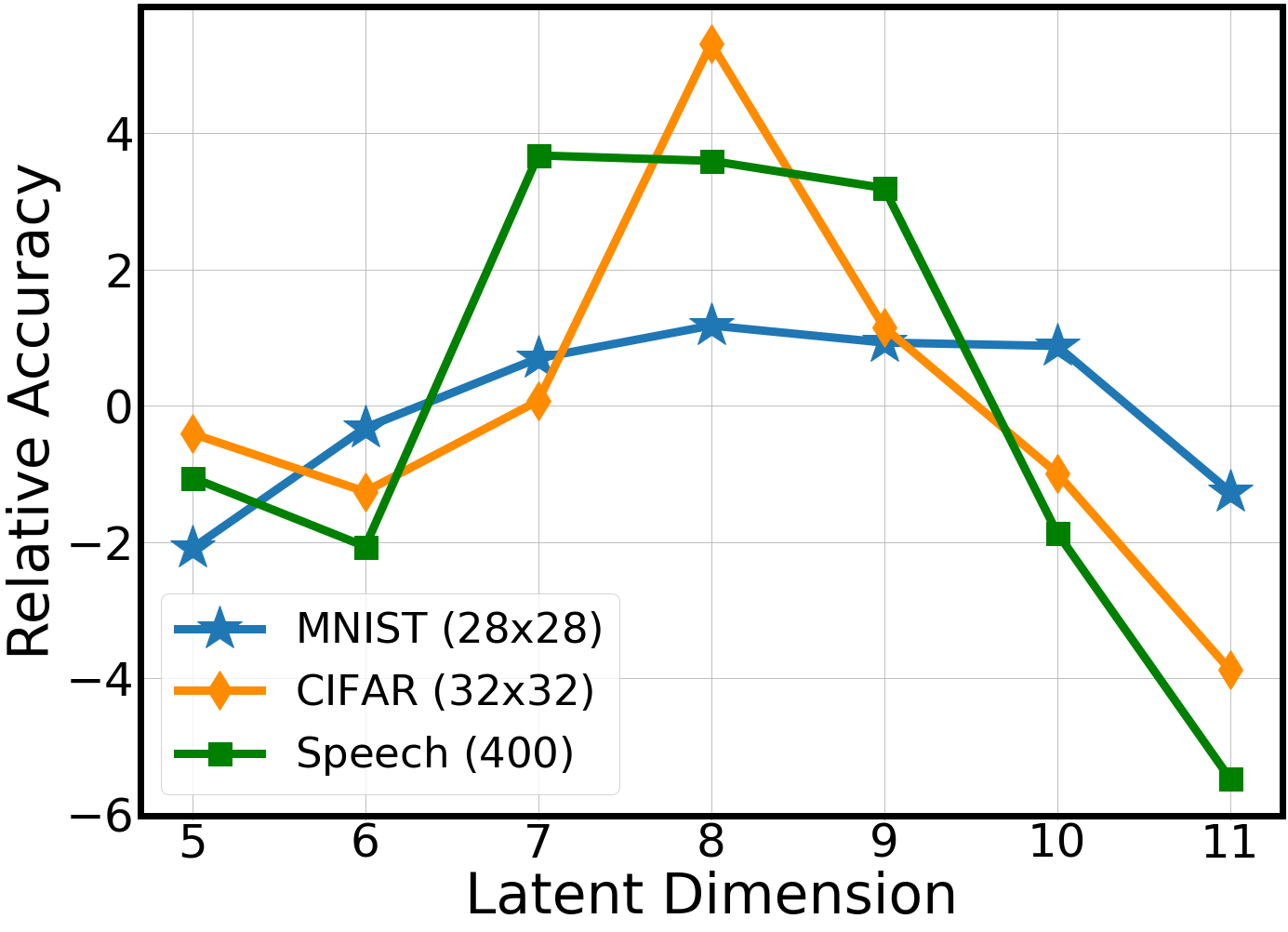}
  \caption{Performance of model as a function of latent representation's size on MNIST ($28 \times 28$), CIFAR-10 ($32 \times 32$), and Speech ($400$) Datasets. The reported performance is normalized with respect to the average performance to ease the comparison between different datasets.}
  \label{fig:size}
\end{figure}

To gain better insights into the effect of the latent dimension on the performance of our models, we ran the experiments and changed the size of the latent representation from $d_h = 32$ to $d_h = 2048$. Figure \ref{fig:size} depicts the average accuracy against the size of the latent layer for MNIST, CIFAR-10, and Fashion MNIST datasets. One interesting observation is that the size of the latent representation does not greatly influence the network's performance \b{on the MNIST, Fashion MNIST and CIFAR-10 datasets}. Even with an overcomplete autoencoder, our network can achieve good results.

An autoencoder learns to extract the most informative features from the input data by passing it through a bottleneck which commonly has fewer dimensions than the original input. In the case of an overcomplete autoencoder, in which the size of the latent representation is equal to or bigger than the dimension of the input data, the network might learn to copy the input to the output without learning any meaningful representation. However, if we regularize the latent representation, for instance, by adding the sparsity constraint, the network can still learn meaningful representation even if the latent size is equal or bigger than the input dimension. The regularization term prevents the network from converging to the trivial solution of copying the input to the output.

\section{Conclusion}
\label{conclusion}

In this paper, we proposed DASVDD, a one-class anomaly detection method that jointly trains an autoencoder and an SVDD on the autoencoder latent representation so that the autoencoder learns to map the normal data to the minimum volume enclosing hypersphere, during the training phase. Previous works have used autoencoders as a means for feature reduction to feed it into a classical one-class classifier such as SVDD.
We defined a customized anomaly score which is a combination of the reconstruction error of the autoencoder and the distance of the lower-dimensional mapping from the hypersphere center. The objective of our model is to minimize this anomaly score on the normal data during training.
We also proposed an effective heuristic for finding a suitable value for the hyperparameter of DASVDD prior to training. Our empirical results on \blue{seven} benchmark datasets have shown that DASVDD outperforms several state-of-the-art anomaly detection algorithms. We also showed that DASVDD does not suffer from problems such as hypersphere collapse that other algorithms may encounter. \blue{It can also be applied to a wide range of input types since it does not rely on any data-specific transformation.} Future studies can explore the efficiency of our method on other domains of application as well as ways to improve its performance by proposing new schemes for hyperparameter tuning and network architecture.


%



\ifCLASSOPTIONcompsoc
  \section*{Acknowledgments}
\else
  \section*{Acknowledgment}
\fi

The authors wish to acknowledge the financial support of the Natural Sciences and Engineering Research Council of Canada (NSERC), and the Fonds de Recherche du Québec - Nature and Technologies (FRQNT).

\ifCLASSOPTIONcaptionsoff
  \newpage
\fi



\bibliographystyle{IEEEtran}
\bibliography{ref}
%



%
\begin{IEEEbiography}[{\includegraphics[width=1.in,height=2in,clip,keepaspectratio]{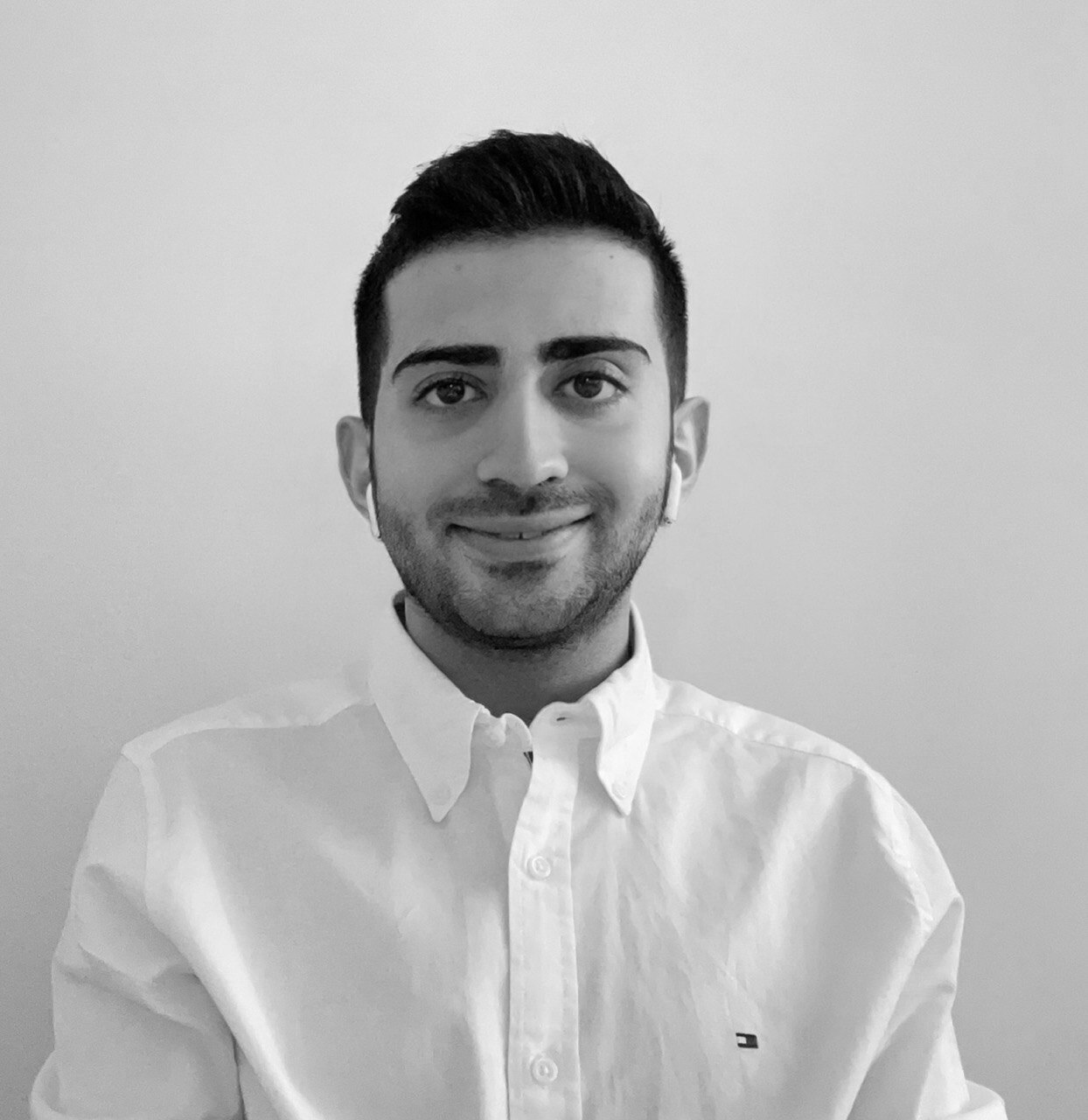}}]{Hadi Hojjati}
is currently a Ph.D. Candidate at the Department of Electrical and Computer Engineering, McGill University. He is also affiliated with Mila Quebec AI-Institute, Montreal, QC, Canada. He received his B.Sc. degree in Electrical Engineering from the Sharif University of Technology, Tehran, Iran, in 2020. His research interest includes representation learning, anomaly detection, and self-supervised learning.
\end{IEEEbiography}
\begin{IEEEbiography}[{\includegraphics[width=1in,height=1.25in,clip,keepaspectratio]{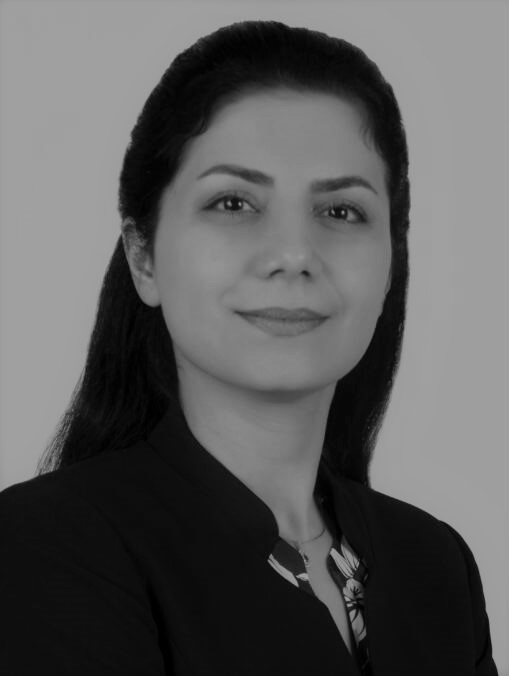}}]{Narges Armanfard}
is currently an Assistant Professor (tenure-track) at the Department of Electrical and Computer Engineering, McGill University and Mila Quebec AI-Institute, Montreal, Quebec, Canada. She received her Ph.D. degree in Electrical and Computer Engineering from McMaster University, Hamilton, ON, Canada, in 2016. She completed her postdoctoral studies at the University of Toronto and University Health Network in 2018. She performs fundamental and applied research in machine learning. Her current research interests include machine learning and related areas in computer vision, reinforcement learning, subspace learning for data clustering, classification, and anomaly detection.
\end{IEEEbiography}








\end{document}